\pdfoutput=1
\documentclass[11pt]{article}
\usepackage{arabtex}
\usepackage[T1]{fontenc}

\usepackage[]{acl}
\usepackage{multirow}
\usepackage{graphicx}
\usepackage{array}
\usepackage{times}
\usepackage{latexsym}
\usepackage[utf8]{inputenc}

\usepackage{utf8}
\usepackage{xspace}

\usepackage{subcaption}
\usepackage{tabularx}
\usepackage{rotating}
\usepackage{amsmath}
\usepackage{tabularx}
\usepackage{nopageno}
\usepackage{amssymb}
\usepackage{newunicodechar}
\newunicodechar{ℝ}{\mathbb{R}}
\usepackage{fixltx2e}
\usepackage[colorinlistoftodos,prependcaption,textsize=tiny]{todonotes}

\setcode{utf8}

\newcommand{\Ar}[1]{{\scriptsize \<#1>\xspace}}

\newcommand{\hadath}{\cal{\textbf{\textit{Wojood\textsuperscript{$Hadath$}}}}\xspace}

\newcommand{\hadathNLI}{\cal{\textbf{\textit{Hadath}}}\textsuperscript{\tiny{NLI}}\xspace}

\newcommand{\testNLI}{\cal{\textbf{\textit{Wojood}}}\textsuperscript{\tiny{OutOfDomain}}\xspace}

\title{Event-Arguments Extraction Corpus and Modeling using BERT for Arabic}

\author{
  \begin{minipage}[t]{0.5\linewidth}
    \centering
    \normalsize Alaa Aljabari \\
    \small \textnormal{Birzeit University \\
    \small \textnormal {aaljabari@birzeit.edu}}
  \end{minipage}
  \hfill
    \begin{minipage}[t]{0.5\linewidth}
    \centering
    \normalsize Lina Duaibes \\
    \small \textnormal{Birzeit University \\
    \small \textnormal{1205358@student.birzeit.edu}}
\end{minipage}
  \\[2.5em] % Increase the vertical space between the rows
  \begin{minipage}[t]{0.5\linewidth}
    \centering
    \normalsize \textbf{Mustafa Jarrar} \\
    \small{Birzeit University \\
    \small{mjarrar@birzeit.edu}}
  \end{minipage}
  \hfill
  \begin{minipage}[t]{0.5\linewidth}
    \centering
    \normalsize \textbf{Mohammed Khalilia}\\
   \small Birzeit University \\
    \small{mkhalilia@birzeit.edu}
  \end{minipage}
}

\begin{document}
\maketitle

\begin{abstract}

Event-argument extraction is a challenging task, particularly in Arabic due to sparse linguistic resources. To fill this gap, we introduce the \hadath corpus ($550$k tokens) as an extension of Wojood, enriched with event-argument annotations. We used three types of event arguments: $agent$, $location$, and $date$, which we annotated as relation types. Our inter-annotator agreement evaluation resulted in  $82.23\%$ $Kappa$ score and $87.2\%$ $F_1$-score. Additionally, we propose a novel method for event relation extraction using BERT, in which we treat the task as text entailment. This method achieves an $F_1$-score of $94.01\%$.
To further evaluate the generalization of our proposed method, we collected and annotated another out-of-domain corpus (about $80$k tokens) called \testNLI and used it as a second test set, on which our approach achieved promising results ($83.59\%$ $F_1$-score). Last but not least, we propose an end-to-end system for event-arguments extraction. This system is implemented as part of SinaTools, and both corpora are publicly available at {\small \url{https://sina.birzeit.edu/wojood}}

\end{abstract}

\section{Introduction}

\begin{figure}
\centering 
 \includegraphics[width=1\columnwidth]{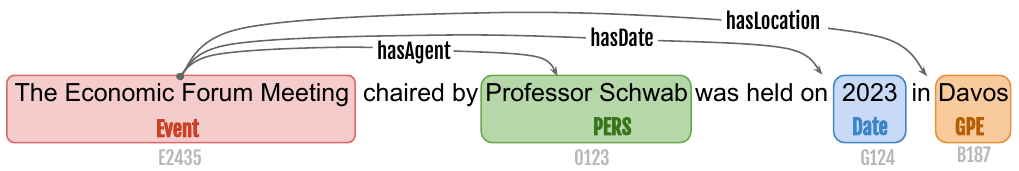}
    \caption{An event annotated with its arguments.}
    \label{fig:introduction}
\end{figure}

%Named Entity Recognition (NER) is pivotal in Natural Language Processing (NLP), as it identifies and classifies named entities, enabling structured information extraction for various applications like knowledge graphs and information retrieval systems \citep{singh2018natural}. 
Understanding and extracting events from text is crucial in natural language understanding \cite{KMSJAEZ24} for applications like disaster monitoring \citep{hernandez2019using}, emergency response \citep{simon2015socializing}, insurance decision support, and fostering community resilience \citep{ahmad2019social}. Events, a type of named entity mentions \cite{JAKBEHO23}, are connected to other entities through their arguments. This connection forms the foundation of the event-argument extraction task, which is closely related to relation extraction. By identifying events and linking them with arguments like agents, locations, and dates, we establish meaningful relationships that enhance applications such as information retrieval systems \citep{singh2018natural}, word sense disambiguation \cite{JMHK23,HJ21b}, and knowledge graph construction \cite{knowdge_graph_2022}. Figure \ref{fig:introduction} shows how the event "The Economic Forum Meeting" is connected with its three arguments.
%Events have been treated as an important type of named entities \citep{araki2018interoperable}.The event-argument extraction task is closely related to the relation extraction task. Recognizing events and connecting them with their arguments, which are also named entities, entails establishing a relationship between these entities. This task is important in various application scenarios, like monitoring of natural disasters \citep{hernandez2019using}, emergency response \citep{simon2015socializing}, aids insurance decisions, and fosters community resilience  \citep{ahmad2019social}.

% https://docs.google.com/presentation/d/151_jFO9wrmgyWVf0_Dh_n5Im7ACVEhVxtAnLppzJvk4/edit?usp=sharing

Despite the significance of event-argument extraction, there is a notable gap in the availability of comprehensive, annotated corpora for this purpose, especially for under-resourced languages like Arabic \cite{DH21,EJHZ22}. To address this gap, we have developed an event-argument extraction corpus specifically for Arabic. We extended the Wojood corpus \cite{JKG22}, which is the largest and most recent NER corpus for Arabic \cite{LJKOA23,JHKTEA24,JAKBEHO23}. We annotated arguments of event entities in Wojood corpus by manually linking each event entity with its corresponding argument entities. As shown in Figure \ref{fig:figures}, three types of arguments are introduced ($agent$, $location$, $date$) for each event. Our Wojood extension (i.e., the event-arguments annotations) is called \hadath.

Furthermore, we introduce a novel method for event-argument extraction using BERT and achieved $94.01\%$ $F_1$-score. Based on \hadath, we generated a dataset of premise-hypotheses sentence pairs (we call it \hadathNLI). We used this dataset to fine-tune BERT, as a natural language inference (NLI) task. To test the generalization of our approach, we have constructed an additional out-of-domain dataset (about $80$k tokens) called \testNLI and used it as a second test set. Our model achieves, again, promising results ($83.59\%$ $F_1$-score). Finally, to streamline the event-argument extraction, we propose an end-to-end system specifically designed for the Arabic event-argument extraction task.
In summary, the contributions of this paper are: 
\begin{enumerate}
    \item \hadath corpus ($550$k tokens) manually annotated with event argument relations. This corpus is used to generate an NLI dataset \hadathNLI ($25$k premise-hypotheses pairs).

    \item \testNLI, an additional annotated corpus ($80$k tokens) for out-of-domain evaluation.
    
    \item Novel methodology for event-argument extraction by framing the task as an NLI problem, achieving high performance.
    
    \item Novel end-to-end system for event-argument relation extraction.

  \end{enumerate}
    
All datasets are available online\footnote{Datasets: {\scriptsize \url{https://sina.birzeit.edu/wojood/}}}, and the end-to-end system is implemented and can be used as part of the SinaTools \cite{HJK24}\footnote{ SinaTools: {\scriptsize \url{https://sina.birzeit.edu/sinatools}}}.

\par This paper is organized as follows: Section \ref{sec:Related Works} reviews previous research. Section \ref{sec:Corpus Annotations} discusses corpus annotations. Section \ref{sec:sec4} presents the inter-annotator agreement analysis. Section \ref{sec:sec5} outlines the methodology. Section \ref{sec:nli_dataset} details the dataset construction, while Section \ref{sec:sec6} covers the experimental setup. Section \ref{sec:out_of_domain_dataset} introduces the out-of-domain dataset. Section \ref{sec:end_to_end} elaborates on the end-to-end system, and Section \ref{sec:ablation_study} explores ablation studies. Finally, Section \ref{sec:conclusion} concludes the paper and discusses future directions.

\section{Related works}

\label{sec:Related Works}
Events are occurrences or actions that happen over time, involving specific participants and locations. They have temporal components and rely on physical entities to take place \cite{J21,JC17}. Extensive research has focused on event extraction and event argument extraction (EAE), especially in English. However, EAE in Arabic is limited, leaving a gap in the literature \cite{chouigui2018tf,alomari2020iktishaf}.

Automated methods, utilizing either statistical algorithms or NLP techniques, are employed to identify relationships between words, using large corpora to detect related terms through their co-occurrence patterns \cite{KAEMKRTM23}. 
For instance, \citet{hkiri2016events} proposed a model to extract event entities from Arabic news articles using the GATE tool, employing a five-stage entity identification process that establishes links between events and their corresponding arguments. However, their dataset is small as it consists of only $1,650$ sentences. 

In a similar context, \citet{AL-Smadi2016} proposed to use an unsupervised rule-based technique to extract events and the relationship with their associated entities from $1,000$ Arabic tweets covering Time, Agent, Location, Target, Trigger, and Product. They linked extracted events and entities to a knowledge base, achieving $75.9\%$ accuracy. This accuracy pertains to the textual representation within the tweet corresponding to the event expression, event type identification ($97.7\%$ accuracy), and event time extraction ($87.5\%$ accuracy). However, despite these achievements, there remains a need for larger and more diverse datasets to further validate these findings.

Multilinguality has increasingly attracted attention across various fields due to its potential to enhance the understanding and processing of diverse linguistic data \cite{JA19, jorgensen-etal-2023-multifin, DJJQQ24}. This direction has notably influenced relation extraction tasks, as seen with SMILER \citep{seganti2021multilingual}, which aims to improve entity and relation extraction, including Arabic. \citet{seganti2021multilingual} fine-tuned HERBERTa on the Arabic subset of SMILER, achieving high performance in identifying relations and entities despite its smaller dataset of $9$k sentences.  This suggests that the unique linguistic features of Arabic may contribute to the model's robustness.
Furthermore, \citet{cabot2023red} introduced valuable resources, namely, SRED\textsuperscript{FM} and RED\textsuperscript{FM}, designed for multilingual relation extraction. SRED\textsuperscript{FM} encompasses over $40$ million triplet instances across $18$ languages, featuring $400$ relation types and $13$ entity types. Their mREBEL model, pre-trained on SRED\textsuperscript{FM}, exhibited a remarkable improvement of  $15$ points in Micro-$F_1$ compared to HERBERTa.  However, their qualitative error analysis excludes Arabic and Chinese because the authors are not proficient in these languages.

Joint extraction models have become increasingly important in NLP. Addressing the need for improved event extraction in Arabic text, \citet{el-khbir-etal-2022-arabie} introduced a joint model for event extraction in Arabic text using the ACE 2005 dataset. They used a graph-based representation to extract entities, relationships, and event triggers, along with their arguments. This approach led to improved Arabic NER accuracy through different experiments and tokenization schemes. 

\par In the realm of biomedicine, \citet{xu2022can} introduced the NBR (NLI improved Biomedical Relation Extraction) method, which verbalizes relations in natural language hypotheses, allowing the model to utilize semantic information effectively in prediction, even with limited data. In contrast, \citet{cao2023zero} addresses cross-lingual EAE challenges through their innovative LanguAge-oriented Prefix-tunIng Network (LAPIN) approach. LAPIN utilizes a language-oriented prefix generator module to handle language variations and a language-agnostic template constructor module to create adaptable templates. Experimental results demonstrate LAPIN's outperforming, achieving an average $F_1$ improvement of $4.8\%$ and $2.3\%$ on two multilingual EAE datasets compared to the previous state-of-the-art models. These efforts collectively contribute to advancing the field of EAE in various languages, including Arabic.

\setcounter{section}{2}
\section{Dataset and annotation}
\label{sec:Corpus Annotations}

\subsection{Corpus Preparation}
To construct an event-argument corpus, we extended the existing  Wojood corpus \citep{jarrar2022wojood}. This is because it is a rich Arabic nested named entity corpus comprising $550$k tokens, supporting $21$ distinct entity types, including categories such as person, organization, location, event, and date. Notably, it incorporates $2,772$ annotated events. According to Wojood guidelines, a mention is annotated as an event if it represents an occurrence of general interest, like battles, wars, sports events, demonstrations, disasters, elections, and national or religious holidays.
%Wojood does not extend to the annotation of relationships between the event mentions and their arguments.

Our objective in this paper is to identify event arguments and establish relationships between these arguments and the respective event entities, as illustrated in Figure \ref{fig:figures}. 

\subsection{Annotation Process}

\begin{figure}[ht!]
\centering

\begin{subfigure}{\columnwidth}
 \centering  
    \includegraphics[width=\columnwidth]{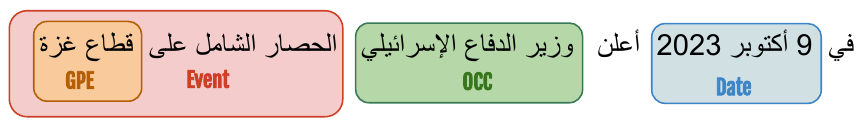}
    \caption{Annotated entities in Wojood.}
    \label{fig:fig1}
\end{subfigure}

\vspace{1em}

\begin{subfigure}{\columnwidth}
    \centering
    \includegraphics[width=\columnwidth]{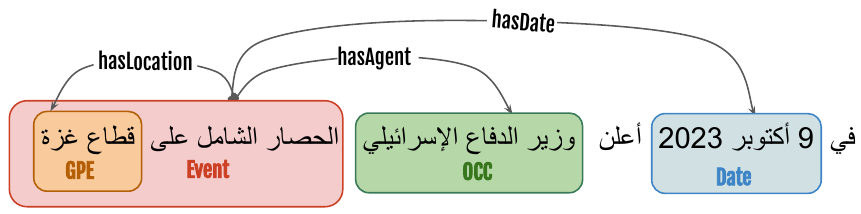}
    \caption{Event with argument entities annotated as relations.}
    \label{fig:fig2}
\end{subfigure}

\vspace{1em}

\begin{subfigure}{\columnwidth}
    \centering
    \includegraphics[width=\columnwidth]{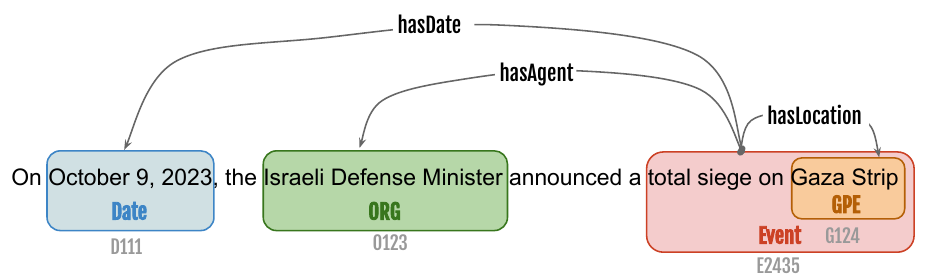}
    \caption{Translation of the above example.}
    \label{fig:third}
\end{subfigure}
        
\caption{Annotating an event entity with its arguments.}
\label{fig:figures}
\end{figure}

Initially, we assigned a unique identifier to each entity in Wojood, see Figure \ref{fig:figures}($a$). Then, we linked these entities with relationships, see Figure \ref{fig:figures}($b$).

% https://docs.google.com/presentation/d/151_jFO9wrmgyWVf0_Dh_n5Im7ACVEhVxtAnLppzJvk4/edit?usp=sharing

\subsubsection{Relationship Types}
\label{sec:relation_types}
We propose to use these relations:
\begin{itemize}
    \item $hasAgent$: specifies participant(s) involved in the event, which can be a PERS, ORG, OCC, or NORP named entities.
    \item $hasLocation$: indicates where the event occurred, which can be GPE, LOC, and FAC named entities.
    \item $hasDate$: points when the event occurred, which can be TIME or DATE. 
\end{itemize}
\par

\subsection{Annotation Guidelines}
\label{sec:sec3.3}
We propose the following guidelines to annotate the corpus:
\begin{enumerate}
    \item  Event arguments are recognized only within the same sentence.
    
    \item Entities with different entity IDs are considered distinct entities. For example, in the sentence (\Ar{مقتل الرئيس المصري أنور السادات} / The killing of the Egyptian president Anwar al-Sadat), the entity (\Ar{الرئيس المصري}/the Egyptian president) and the entity (\Ar{أنور السادات} / Anwar al-Sadat) are regarded as separate entities, thus two agents in the event. However, in reality, they refer to the same individual.
    
    \item The same event may have multiple agents, as in (\Ar{ توقيع اتفاقية تعاون بين الحكومة اللبنانية و البنك المركزي  } / Signing a cooperation agreement between the Lebanese government and the Central Bank). In Figure \ref{fig:fig3}, (\Ar{الحكومة اللبنانية} / the Lebanese government) and (\Ar{البنك المركزي} / the Central Bank) are both agents of the same event.
    
    \item In the same sentence, two different event entities can share the same argument. For example, in the sentence (\Ar{الوضع السياسي متوتر في مصر بعد حرب عام 1967 (النكسة)}/The political situation in Egypt is tense after the 1967 War (Al Naksa)), there are two event entities, according to Wojood guidelines. The first event is (\Ar{حرب عام 1967}/The 1967 war), and the second event is (\Ar{(النكسة)}/Al Naksa). In this way, the entity (\Ar{عام 1967}/the year 1967) is considered the argument for both events in the sentence.

\begin{figure}[ht]
\centering
   \includegraphics[scale=0.7]{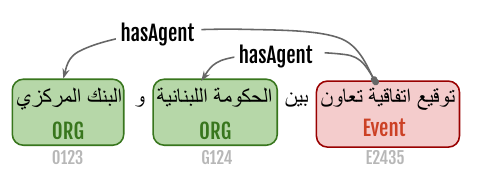}
    \caption{An event with more than one agent}
    \label{fig:fig3}
\end{figure}

% https://docs.google.com/presentation/d/151_jFO9wrmgyWVf0_Dh_n5Im7ACVEhVxtAnLppzJvk4/edit?usp=sharing
\end{enumerate}

\subsection{Corpus Statistics}
Wojood comprises $550$k tokens with $2,772$ event entities.
Among these entities, we annotated $1,974$ events with event-arguments relations - the other $798$ do not have arguments.  Notably, there are $355$ ($18\%$) events that are annotated with at least two arguments, and $77$ ($18\%$ of agent relations) annotated with multiple agents. This underscores the corpus’s rich and diverse interconnection of events and their roles.
The number of instances for each relation type is shown in Table \ref{tbl:wojood_konooz_event}.
\begin{table}[ht!]
\small
\centering
\begin{tabular}{@{}lc@{}}%{p{1.2cm}p{2cm}p{1.5cm}p{1.5cm}}
\hline
    %\multirow{2}{*}{ Type} & \multirow{2}{*}{Name} & \multicolumn{2}{c}{} \\
    Relation & Count
     \\
\hline
 $hasAgent$ & 423\\
 $hasLocation$ & 833\\
 $hasDate$ & 1,332\\
\hline
\textbf{Total} & \textbf{2588}
\end{tabular}
\caption{\label{tbl:wojood_konooz_event}
Number of relations in \hadath }
\end{table}

\section{Inter-annotator Agreement}
\label{sec:sec4}

To evaluate the quality of our annotations, we randomly selected $5\%$ of the annotations and asked our two annotators to annotate in parallel. We computed the Inter-annotator Agreement (IAA) using Cohen's $kappa$ and $F_1$-score. The results in Table \ref{tab:tab4} illustrate high agreement. 

\begin{table*}[ht]
\small
  \centering
\begin{tabular}{ *{6}{l} }
\hline
  \textbf{Relation} & \textbf{TP}  & \textbf{FN} & \textbf{FP}  &  $\mathbf{\kappa}$  &   \textbf {F1-Score}  \\ \hline  

  $hasAgent$ &  37 &  10 &  10 &  67.85\% & 79\% 
\\ 
  
$hasLocation$  &  29 &  2 &  2 &  91.70\% &  94.00\% 
\\ 

$hasDate$ &  43 &  2 &  6 &  87.15\% &  91\% 
\\ \hline

 \textbf{Overall} & \textbf{109 \scriptsize count}  & \textbf{14 \scriptsize count} & \textbf{18 \scriptsize count} & \textbf{82.23\% \scriptsize macro} & \textbf{87.20\% \scriptsize micro}\\ \hline
\end{tabular}
\caption{Overall IAA for each relation.\label{tab:tab4} }
\end{table*} 

%Ensuring annotation quality is crucial, with inter-annotator agreement (IAA) serving as a key metric to quantifies the level of agreement among annotators. Following \citet{ning2018multi} and \citet{ikuta2014challenges}, our second annotator sampled 5\% of the annotated data from the Wojood corpus for IAA calculations. Results are detailed in Table \ref{tab:tab4}.

%\subsection{Calculations}
%IAA is typically assessed using Kappa, which we employed for event relation and argument annotation. We used two evaluation metrics for IAA: Kappa ($\kappa$) and F1-score.

\subsection{Calculating $kappa$ }
We assessed annotator agreement for three relations. Agreement happens when both annotators assign the same relationship type. The $kappa$ coefficient \citep{eugenio2004kappa} is defined as:

 \begin{equation}
 \small
    K = \frac{P_0-P_e}{1-P_e}
\end{equation} 
where $P_o$ is the observed agreement, and $P_e$ is the expected agreement. $P_e$ is calculated as :
\begin{equation}
\small
    P_e = \frac{1}{N^2}\sum_{T}n_{T1} \times
    n_{T2}
\end{equation}
\subsection{Calculating $F_1$-score}
The $F_1$-score for a specific relation (e.g., $hasAgent$) is calculated using Equation \ref{eq:f1}. True positives (TP) are those when annotators agree, while false negatives (FN) and false positives (FP) arise from disagreements. FN arises from the first annotator's disagreement, while FP arises from the second annotator's disagreement.
\begin{equation}
\small
    F_1-Score = \frac {2TP} {2TP+FN+FP}
\label{eq:f1}
\end{equation}

\subsection{Discussion and Annotation Challenges} 
\label{sec:sec3.5}
We encountered several challenges, including: 
\begin{enumerate}
\item Although event arguments should be encompassed within the sentence referencing the event, it was not always easy for the annotators to decide if entities within the same sentence as the event should be annotated as event entities.

\item 
In the $hasAgent$ relation, it can sometimes be challenging to determine whether an entity serves as an agent for the event. See the example in Figure \ref{fig:fig4}.

\begin{figure}[h!]
\centering
   \includegraphics[scale=0.7]{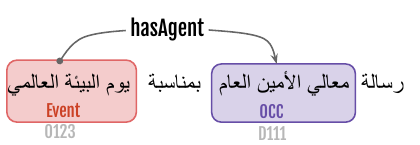}
\caption{Example of disagreement: whether the "Secretary-General" is an agent in the sentence (\textit{letter by the Secretary-General on the occasion of World Environment Day}).}
    \label{fig:fig4}
\end{figure}
% https://docs.google.com/presentation/d/151_jFO9wrmgyWVf0_Dh_n5Im7ACVEhVxtAnLppzJvk4/edit?usp=sharing
\item The annotators have different levels of experience; one is skilled at defining event arguments, while the other is not. Consequently, one annotator agreed to designate a specific entity as the agent, while the second did not.
\end{enumerate}

\section{Even-Argument Extraction (EAE)} 
\label{sec:sec5}
\begin{figure*}
%https://docs.google.com/presentation/d/1d2FhCTkPd5nGBckcXfBiRx-8IJnwvRXg7d7XPn4iE8o/edit#slide=id.p
\centering
  \includegraphics[scale=0.65]{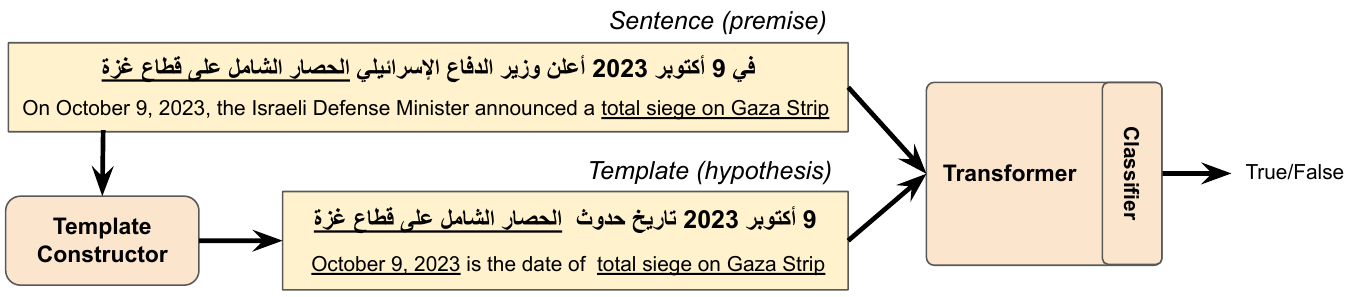}
  \caption{Framing the EAE task as NLI task.}
      \label{fig:methodology}
\end{figure*}

In this section, we provide an in-depth explanation of our approach to addressing the EAE task through the NLI task. Figure \ref{fig:methodology} illustrates our framing of the EAE problem.

% https://docs.google.com/presentation/d/151_jFO9wrmgyWVf0_Dh_n5Im7ACVEhVxtAnLppzJvk4/edit?usp=sharing

\subsection{Problem Formulation}
The objective of the EAE task is to identify the relationships between the event(s) in a sentence and the named entities mentioned in the same sentence. That is, given a sentence $s$ annotated with a set of event entities $E=\{e_i\}_{i=1}^n$ and a set of other named entities $N=\{n_j\}_{j=1}^m$, the goal of EAE is to identify the relation $r\in R$ for each pair $(e_i,n_j)$ in $s$, where $R\in \left \{ hasAgent, hasLocation, hasDate\right \}$. 
%Each annotated sentence $s$ in the dataset contains a set of event mentions $E=\{e_1, ...,e_i\}$ and named entities $N=\{n_1, .., n_j\}$. Our approach is designed to extract the relation $r_{i}\in R$ between $e_i$ and $n_j$, where $R\in \left \{ hasAgent, hasLocation, hasDate\right \}$ $\cup \left \{ noRelation \right \}$.

\subsection{Event Relation Extraction as NLI}

We propose to solve the EAE by framing it as a Natural Language Inference (NLI) task. 
In NLI, we assess whether one sentence (the \textit{premise}) entails another (the \textit{hypothesis}).
That is, a pair of sentences is classified as $True$ or $False$. 
To extract an event argument relation from a sentence (See Figure \ref{fig:methodology}), we treat the original sentence as a premise and generate the hypothesis automatically. The hypothesis is another sentence generated using a template, to represent a possible relationship. 
%Following \citep{nli_2023}, we reformulate EAE as an NLI task. In NLI, the model is used to evaluate the relationship between two texts: the premise and the hypothesis. The premise is a sentence that offers context for reasoning, while the hypothesis is a claim based on the premise. The objective is to determine whether the hypothesis entails, contradicts, or has no relationship (neutral) with the premise. % The goal is to determine if the hypothesis logically follows (entailment), contradicts (contradiction), or has no relationship (neutral) with the premise.  
In other words, we propose to treat EAE as a binary NLI task focusing on entailment. The input sentence $s$ is the premise, while the hypothesis is a verbalized template representing a relation $r$ between event $e_i$ and a named entity $n_j$  mentioned in the same sentence. The model then determines if the premise "entails" the hypothesis (as classification $True$/$False$), indicating the existence of the relation $r$ between $e_i$ and $n_j$.

 %See the example in Figure \ref{fig:NLI_example}.

\subsubsection*{Template Construction}

\label{sec:template_construction}

For each of the three relations, we designed a template parameterized over the named entities mentioned within the same sentence. Each template has two placeholders: Event Placeholder $P_{event}$ and Entity Placeholder $P_{entity}$. These placeholders are filled with event mention $e_i$ and named entity mention $n_j$, respectively.
Figure \ref{fig:methodology} illustrates how a template can be used to generate a hypothesis for the $hasDate$ relation between the (\Ar{الحصار الشامل على قطاع غزة} / Total siege on Gaza strip) event and the
(\Ar{9 أكتوبر 2023} /October 9, 2023) entity. The three templates used are shown in Table \ref{tab:relation-templates} 

\begin{table*}[htbp]
\centering
\small
\begin{tabular}{@{}p{0.15\linewidth}p{0.25\linewidth}p{0.55\linewidth}@{}}
\hline
\textbf{Relationship} & \textbf{Template} & \textbf{Example} \\
\hline
$hasAgent$ & \small \textcolor{blue}{$P_{event}$} \tiny\< أحد الفاعلين في> \small \textcolor{red}{$P_{entity}$} & \textcolor{blue}{\tiny \<الحصار الشامل على قطاع غزة>} \tiny\<أحد الفاعلين في> \textcolor{red}{\tiny\<وزير الدفاع الإسرائيلي>} \\
& & \small  \textcolor{red}{The Israeli defense minister} is an agent in \textcolor{blue}{Total siege on Gaza strip} \\
\hline
$hasLocation$ & \small \textcolor{blue}{$P_{event}$} \tiny\<مكان حدوث> \small \textcolor{red}{$P_{entity}$} & \small \textcolor{blue}{\tiny\<الحصار الشامل على قطاع غزة>} \tiny\<مكان حدوث>  
 \textcolor{red}{\tiny\<قطاع غزة>} \\
& & \small \textcolor{red}{Gaza strip} is place of occurring \textcolor{blue}{Total siege on Gaza strip} \\
\hline
$hasDate$ & \small  \textcolor{blue}{$P_{event}$} \tiny\<تاريخ حدوث> \small \textcolor{red}{$P_{entity}$} & \textcolor{blue}{\tiny\<الحصار الشامل على قطاع غزة>} \tiny\<تاريخ حدوث> \textcolor{red}{\tiny\<9 أكتوبر 2023>} \\
& & \small \textcolor{red}{October 9,2023} is the date of occurring \textcolor{blue}{Total siege on Gaza strip} \\
\hline
\end{tabular}
\caption{Templates utilized during the testing phase.}
\label{tab:relation-templates}
\end{table*}

%The same templates serve for both positive and negative instances. In a negative instance, $e$ represents any other named entity present in the sentence $x$ which holds no association with the event.

\subsubsection*{Sentence Encoder}

The sentence encoder is used to extract representations for input text. In our approach, the input text is a pair consisting of a premise (the input sentence $s$) and a hypothesis (the filled-in template $h$).  A transformer encoder, denoted by $T$, is used to process the input sequence and derive its representation, $T \rightarrow \mathbb{R}^{d}$. This encoder effectively captures contextual information and semantic relationships within the text pair generating representation $H$ for the sentence pairs. Formally,

\begin{equation}
\small
    H=T\left ( [CLS] \textbf{s} [SEP] \textbf{h} \right )
\end{equation}

where $[CLS]$ is the special classification token, $[SEP]$ is a separator between the premise and hypothesis, and $d$ is the encoding dimension.

\subsubsection*{Relation Classifier}

The feature vector $H$ from the sentence encoder is input into a fully connected layer. This layer calculates the predicted probability that the premise entails the hypothesis, with a $True$ result indicating that the relationship $r$ specified in $h$ exists between the event $e$ and the entity $n$. Formally,
\begin{equation}
\small
    \hat{y}_i=\sigma \left ( H W + b \right )
\end{equation}
where $\hat{y}$ indicates whether the hypothesis holds a positive relation, while $W$ and $b$ represent a weight matrix and a bias term, respectively.
 %The main goal during training, across all instances and relations in our dataset, is to minimize the cross-entropy loss $L_{CE}$ between the true relation label distribution and the predicted label distribution across all model parameters.

\subsubsection*{Training Objective}

Our training objective is to prioritize the accurate identification of positive instances over negatives. To accomplish this, we utilize a weighted cross-entropy loss $L_{WCE}$ (Eq. \ref{eq:WCE}), which penalizes misclassifications of positive instances more heavily. Additionally, we use Noise Contrastive Estimation Loss $L_{NCE}$ (Eq. \ref{eq:NCE})  to improve the discrimination between positive and negative instances \cite{contrastive_learning_2020}. The final loss function is expressed as $Loss = L_{WCE} + L_{NCE}$.
%L_{WCE} = -\frac{1}{N} \sum_{i=1}^{N} \bigg ( w_{p}  y_i  \log(\hat{y_i}) + w_{n}  (1 - y_i) \log(1 - \hat{y_i}) \bigg )
%Our training objective addresses the challenge of imbalanced data (See Section \ref{sec:NLI dataset}) by employing a weighted cross-entropy loss $L_{WCE}$ (Eq. \ref{eq:WCE}). This assigns higher penalties to misclassifications of positive instances, prioritizing correct identification despite their scarcity. We also utilize Noise Contrastive Estimation Loss $L_{NCE}$ (Eq. \ref{eq:NCE}) to emphasize the importance of correctly identifying positive instances, and enhancing discriminative capabilities. The final loss function is $Loss = L_{WCE} + L_{NCE}$.
%L_{WCE} = -\frac{1}{N} \sum_{i=1}^{N} \bigg ( w_{p}  y_i  \log(\hat{y_i}) + w_{n}  (1 - y_i) \log(1 - \hat{y_i}) \bigg )
\begin{equation} 
\small
\begin{split}
L_{WCE} &= -\frac{1}{N} \sum_{i=1}^{N} \bigg( w_{p} y_i \log(\hat{y_i}) \\
&\quad+ w_{n} (1 - y_i) \log(1 - \hat{y_i}) \bigg)
\end{split}
\label{eq:WCE}
\end{equation}

\begin{equation} 
\small
L_{NCE} = -\frac{1}{N} \sum_{i=1}^{N} \log \left( \frac{\exp(s(y_i) / \tau)}{\sum_{j} \exp(s(y_j) / \tau)} \right)
\label{eq:NCE}
\end{equation}

%L_{NCE} = - \frac{1}{N} \sum_{i=1}^{N} \log \left( \frac{\exp(\hat{y}_{i, y_i} / \tau)}{\exp(\hat{y}_{i, y_i} / \tau) + \sum_{j \neq y_i} \exp(\hat{y}_{i, j} / \tau)} \right)
%L_{NCE} = - \frac{1}{N} \sum_{i=1}^{N} \log \left( \frac{\exp(s(y_i) / \tau)}{\exp(s(y_i) / \tau) + \sum_{j \neq y_i} \exp(s(y_j) / \tau)} \right)

% $\hat{y}_{i, y_i}$ is the score/logit of the true class for sample i.
% \hat{y}_{i, j} is the score/logit of class j for sample.

\section{Construct \hadathNLI  Dataset}
\label{sec:nli_dataset}
Based on our annotated \hadath corpus, we created an NLI dataset (called \hadathNLI), which is a dataset of premise-hypotheses sentence pairs that we use to train the EAE model. 
% (4) Pairing premise sentences with corresponding hypothesis sentences
To create this dataset, we carried out the following  steps:  

\subsubsection*{1. Premise Sentences Preparation}
We extracted sentences from the \hadath corpus and utilized them as premises. These sentences were then split into training ($70$\%) and testing ($30$\%) sets. This split is important to avoid overlap between premises in the two splits during the generation of positive and negative pairs.

\subsubsection*{2. Hypothesis Sentences Preparation}

Multiple hypotheses are generated for each premise sentence, utilizing the entities mentioned within the sentence. These entities are annotated as arguments of a specific event in the \hadath corpus. That is, based on the event-argument relation annotations (see section \ref{sec:relation_types}), each event and its arguments are used to generate a hypothesis using the template for relation $r$. This hypothesis is then paired with the premise sentence to form a \textbf{positive} pair.
%For the training phase, four templates are designed for each relation, each with a particular verbalizer (see Table \ref{tab:relation-templates_train}). This step is for data augmentation and enhancing the contextual diversity of the training set.  For the test phase, one template per relation was selected from the training templates to generate the hypothesis (see Table \ref{tab:relation-templates}). 

%each paired with the premise sentence from the training split.
\subsubsection*{3. Generating Negative Pairs}
In premise sentences, entities that are not linked with events (i.e., not event arguments) are used to generate \textbf{negative} pairs.
%That is, to generate negative instances for each relation, negative entities in the premise sentence are categorized, as described in Section \ref{sec:relation_types}, into $hasAgent$, $hasLocation$, or $hadDate$ negative arguments for the target event. Subsequently, each entity is utilized to generate a negative instance using its corresponding relation template.
%For example, negative arguments labeled as 'PERS' are used to generate negative instances for the $hasAgent$ relation. This systematic categorization enhances the model's ability to effectively differentiate between ambiguous arguments.
That is, we link events with entities that are not their correct arguments (i.e., not annotated as a relation in \hadath), and generate them as hypotheses using templates. Each generated hypothesis is paired with its premise and used as a negative pair.

\subsection*{\hadathNLI Dataset Statistics}
For the training phase, four templates are designed for each relation, each with a particular verbalizer (see Table \ref{tab:relation-templates_train}). This step aims at data augmentation and enhancing the contextual diversity of the training set.  For the test phase, one template per relation was selected from the training templates to generate the hypothesis (see Table \ref{tab:relation-templates}). 
The final dataset consists of $25,473$ pairs, comprising $10,478$ positive and $14,995$ negative pairs. Table \ref{table:training_statistics} presents more statistics.

\begin{table}[h!]
\centering
\small
\begin{tabular}{|p{0.1\linewidth}|p{0.21\linewidth}rr|r|}
\hline
\textbf{Phase} & \textbf{Pairs} & \textbf{Positive} & \textbf{Negative} & \textbf{Total} \\
\hline
\multirow{3}{*}{\begin{turn}{270} \centering Train  \end{turn}}
& $hasAgent$ & 1,248 & 6,156 & \textbf{7,404} \\
& $hasLocation$ & 2,268 & 4,456 & \textbf{6,724} \\
& $hasDate$ & 3,716 & 2,948 & \textbf{6,664} \\
\hline
&\textbf{SubTotal} & \textbf{7,232} & \textbf{13,560} & \textbf{20,792} \\
\hline
\multirow{3}{*}{\begin{turn}{270} \centering Test  \end{turn}}
& $hasAgent$ & 111 & 653 & \textbf{764} \\
& $hasLocation$ & 267 & 464 & \textbf{728} \\
& $hasDate$ & 403 & 318 & \textbf{721} \\
\hline
&\textbf{SubTotal} & \textbf{778} & \textbf{1,435} & \textbf{2,213} \\
\hline
\hline
&\textbf{Total} & \textbf{8,010} & \textbf{14,995} & \textbf{23,005} \\
\hline
\end{tabular}
\caption{Number of pairs in the \hadathNLI Dataset}
\label{table:training_statistics}
\end{table}

\begin{table*}[ht]
\centering
\small
\begin{tabular} %{clll}
{@{}p{0.07\linewidth}p{0.3\linewidth}p{0.26\linewidth}p{0.30\linewidth}@{}}
\hline
\textbf{Template} & \textbf{$hasLocation$} & \textbf{$hasAgent$} & \textbf{$hasDate$} \\
\hline
\normalsize $t_1$ &\small \textcolor{blue}{$P_{event}$}   \tiny (site of) \<هو موقع> \small \textcolor{red}{$P_{entity}$} &  \small \textcolor{blue}{$P_{event}$} \tiny (actor of)\<أحد المتأثرين في>  \small \textcolor{red}{$P_{entity}$} & \small \textcolor{blue}{$P_{event}$} \tiny (time of) \<هو زمن> \small \textcolor{red}{$P_{entity}$} \\
%& \tiny\<هو موقع>: is the site of & \tiny\<أحد الفاعلين في>: one of the actors& \tiny\<هو زمن>: is time of\\

\hline
\normalsize $t_2$ & \small \textcolor{blue}{$P_{event}$} \tiny (place of occurring) \<مكان حدوث> \small \textcolor{red}{$P_{entity}$} &  \small \textcolor{blue}{$P_{event}$} \tiny (agent in) \<أحد الفاعلين> \small  \textcolor{red}{$P_{entity}$} & \small \textcolor{blue}{$P_{event}$} \tiny (date of occurring) \<تاريخ حدوث> \small \textcolor{red}{$P_{entity}$} \\
%& \tiny \<مكان حدوث> : place of occurring & \tiny \<أحد المتأثرين>: affected by & \tiny \<تاريخ حدوث>: date of occurring\\

\hline
\normalsize$t_3$ & \small\textcolor{red}{$P_{entity}$} \tiny (in) \<في> \small \textcolor{blue}{$P_{event}$} & \small \textcolor{red}{$P_{entity}$} \tiny (related with)\<له علاقة مع> \small \textcolor{blue}{$P_{event}$} & \small \textcolor{red}{$P_{entity}$} \tiny (happened at) \<حدث في>  \small \textcolor{blue}{$P_{event}$} \\
%& \tiny\<في> : in & \tiny\<له علاقة مع>: related with & \tiny\<حدث في>: happened at\\
\hline

\normalsize $t_4$ & \small \textcolor{red}{$P_{entity}$} \tiny (in)\<في>  \small \textcolor{blue}{$P_{event}$} \tiny (happened) \<وقع >
& \small \textcolor{blue}{$P_{event}$}  \tiny (in) \<في> \small \textcolor{red}{$P_{entity}$} \tiny (participate) \<شارك> 
& \small \textcolor{red}{$P_{entity}$}  \tiny (at date) \<بتاريخ> \small \textcolor{blue}{$P_{event}$} \tiny (happened) \<حدث>\\
%& \tiny \<وقع .. في> : happened .. in & \tiny \<شارك .. في>: participate .. in & \tiny\<حدث .. بتاريخ>: happened at .. date\\
\hline
\end{tabular}
\caption{Templates utilized in the training phase. Template Set $t_2$ is selected for the testing phase.}
\label{tab:relation-templates_train}
\end{table*}

\section{EAE Modeling Experiments}
\label{sec:sec6}

 %The F1 score is employed as the evaluation metric for relation classification.
The \hadathNLI is used to train an EAE model.

\subsection{Training Hyperparameters}
%During training, we employed a K-fold strategy with $k=5$, partitioning our dataset into $70\%$ training and development sets and $30\%$ testing sets. The \hadath sentences are carefully split before generating positive and negative instances to prevent information sharing between training and testing, thereby ensuring the robustness of model evaluation.

During training, we employed a $k$-fold strategy with $k=5$ to ensure the robustness of model evaluation. For all experiments, we utilized the \texttt{UBC-NLP/ARBERTv2} \cite{arbert_2021} as the text encoder model, with a learning rate of {$2e^{-5}$} and the $AdamW$ optimizer with a weight decay of $1e^{-8}$. In the $L_{WCE}$ loss function, $w_{pos}=1$ and $w_{neg}=.5$, while $\tau = 1$ in $L_{NCE}$ loss function.

%In $L_{WCE}$ loss function, we use $w_{pos} = 1$ and $w_{neg} = 0.2$. Additionally, in $L_{NCE}$, we set $\tau = 1$.

\subsection{Results and Discussion}

Table \ref{table:results} shows the obtained results using the aforementioned experimental setups. We average the scores across five folds and report the model with the best average $F1 score$. Table \ref{table:EAE_results} provides insight into the NLI model performance in each relation. Note that this NLI performance represents the accuracy of our model in classifying sentence pairs, which is not the EAE accuracy. The accuracy of event-argument relation extraction (EAE) shall be presented in section \ref{sec:end_to_end}.

Notably, our NLI model achieves high $F1 scores$ across all event relations, with the $hasDate$ relation exhibiting the highest score $(96.44\%)$. 
%This indicates the model's proficiency in identifying event arguments associated with temporal occurrences. 
Additionally, the negative relations also achieved a notably high $F_1$ score of $95.29\%$, indicating the model's effectiveness in recognizing entities unrelated to the event.

%Although the performance slightly decreases for the $hasAgent$ relations, the model continues to effectively extract the $hasAgent$ relation, handling challenges such as entity ambiguity and syntactic complexities with strong capabilities. 

To validate this remarkable performance of the model and to test its generalization, we constructed a new corpus and conducted additional out-of-domain experiments. 

\begin{table}[h!]
\small
\centering
\begin{tabular}{lrrrr}
\hline
\textbf{Class} & \textbf{Support} & \textbf{$P$} & \textbf{$R$} & \textbf{$F_1$} \\
\hline
Positive& 778 & 90.06 & 92.42 &92.24 \\
Negative &1435 &  95.99 & 95.68 & 95.78  \\
Average & &  &  &\textbf{94.01}  \\
\hline
\end{tabular}
\caption{ \label{table:results} Results on the \hadathNLI Test Set.}
\end{table}

\begin{table}[h!]
\small
\centering
\begin{tabular}{lrrrr}
\hline
\textbf{Relation} & \textbf{Support} & \textbf{$P$} & \textbf{$R$} & \textbf{$F_1$}  \\
\hline
$hasAgent$  &111 &  82.60 & 85.58 & 84.07\\
$hasLocation$ & 267& 92.40 & 85.70 &89.88  \\
$hasDate$ & 403& 95.38 & 97.51 & 96.44  \\
\hline
\end{tabular}
\caption{ \label{table:EAE_results} Results on the \hadathNLI Test Set, per relation}
\end{table}

\section{Additional \testNLI Dataset}
\label{sec:out_of_domain_dataset}
To evaluate the model's generalization and its robustness to contexts beyond the \hadath domains, we constructed a new out-of-domain corpus (referred to as the \testNLI corpus). We then extracted an NLI dataset and used it for model testing.

\subsection{Corpus Preparation}
This corpus covers $10$ distinct domains: economics, finance, politics, science, technology, art, law, agriculture, history, and sports each containing nearly $8$k tokens. The corpus covers events from 2010 to 2022, manually collected from news websites such as Aljazeera, and Alarabiya, totaling $80k$ tokens. 

We used the same Wojood annotation guidelines to maintain consistency.  Table \ref{tbl:test-data} shows the number of instances for each relationship type.

\begin{table}[ht!]
\small
\centering
\begin{tabular}{@{}lc@{}}%{p{1.2cm}p{2cm}p{1.5cm}p{1.5cm}}
\hline
    %\multirow{2}{*}{ Type} & \multirow{2}{*}{Name} & \multicolumn{2}{c}{} \\
    Relation & Count
     \\
\hline
 $hasAgent$ & 138\\
 $hasLocation$ & 218\\
 $hasDate$ & 125\\
\hline
\textbf{Total} & \textbf{481}
\end{tabular}
\caption{\label{tbl:test-data}
Number of relations in \testNLI }
\end{table}

\subsection{Construct \testNLI Dataset}

An NLI dataset was generated from \testNLI using the same methodology as \hadathNLI, employing only the templates designated for testing, resulting in a total of $1124$ pairs. Detailed statistics are provided in Table \ref{table:test_nli_statistics}.

\iffalse
\begin{table}[h!]
\centering
\small
\begin{tabular}{|p{0.21\linewidth}rr|r|}
\hline
 \textbf{Pairs} & \textbf{Positive} & \textbf{Negative} & \textbf{Total} \\
\hline

 $hasAgent$ & 69 & 224 & \textbf{293} \\
 $hasLocation$ & 174 & 169 & \textbf{343} \\
 $hasDate$ & 110 & 151 & \textbf{216} \\
\hline
\textbf{Total} & \textbf{353} & \textbf{544} & \textbf{897} \\
\hline
\end{tabular}
\caption{Pairs in the \testNLI Dataset are used testing only}
\label{table:test_nli_statistics}
\end{table} 
\fi

\begin{table}[h!]
\centering
\small
\begin{tabular}{|p{0.21\linewidth}rr|r|}
\hline
 \textbf{Pairs} & \textbf{Positive} & \textbf{Negative} & \textbf{Total} \\
\hline

 $hasAgent$ & 108 & 304 & \textbf{412} \\
 $hasLocation$ & 201 & 207 & \textbf{408} \\
 $hasDate$ & 124 & 180 & \textbf{304} \\
\hline
\textbf{Total} & \textbf{433} & \textbf{691} & \textbf{1124} \\
\hline
\end{tabular}
\caption{NLI datasets based on \testNLI}
\label{table:test_nli_statistics}
\end{table}
\subsection{Experiments and Results}

The \testNLI is used to evaluate the EAE model, which was trained on the \hadathNLI. The results, presented in Table \ref{table:results_test_nli}, highlight the model's robust generalization across diverse domains, despite encountering challenges like domain-specific vocabulary and linguistic nuances. Despite a slight performance decline compared to the \hadathNLI test set, the model maintained a high overall average $F_1-score$ of $83.38\%$ on the \testNLI.

\begin{table}[h!]
\small
\centering
\begin{tabular}{lrrrr}
\hline
\textbf{Class} & \textbf{Support} & \textbf{$P$} & \textbf{$R$} & \textbf{$F1$} \\
\hline
Positive & 478& 71.05 & 78.03 &74.38  \\
Negative&1809 &  94.04 & 91.60 & 92.80 \\
Average & &  & & \textbf{83.59} \\
\hline
\end{tabular}
\caption{ \label{table:results_test_nli} Experimental Results on \testNLI.}
\end{table}

\iffalse
\begin{table}[h!]
\small
\centering
\begin{tabular}{lrrrr}
\hline
\textbf{Class} & \textbf{Support} & \textbf{$P$} & \textbf{$R$} & \textbf{$F1$} \\
\hline
Positive & 387& 69.87 & 84.50 &76.49  \\
Negative&1494 &  95.75 & 90.56 & 93.09  \\
Average & &  & & \textbf{84.79} \\
\hline
\end{tabular}
\caption{ \label{table:results_test_nli} Experimental Results (F1-score \%) on \testNLI.}
\end{table}
\fi
%This cross-dataset experiment ensures that our model's performance is not limited to the training dataset and can effectively generalize to unseen data.

%For both datasets, we utilized the F1 score as the evaluation metric for relation classification.
\begin{figure*}
    \centering
    \includegraphics [scale=0.75]{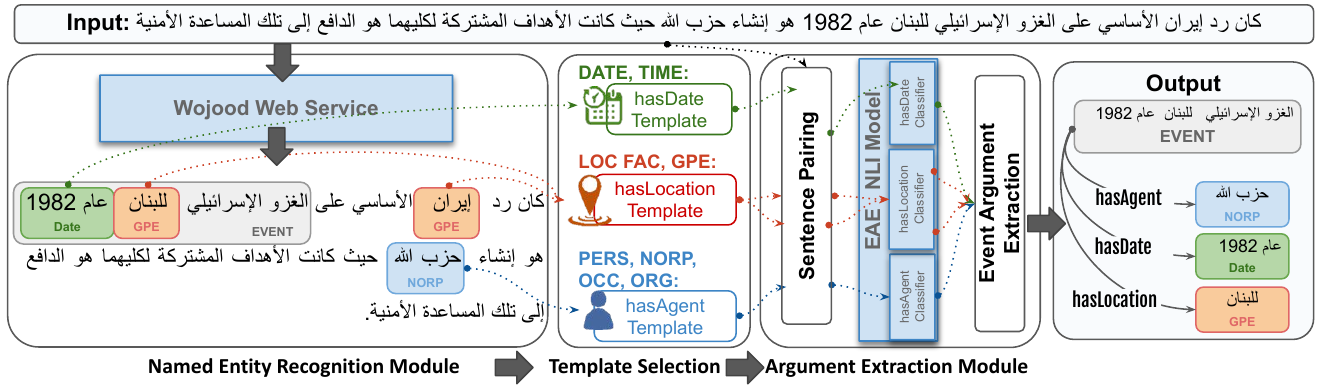}
    \caption{End-To-End Event Argument Extraction Architecture.}
    \label{fig:inference}
\end{figure*}
% https://docs.google.com/presentation/d/151_jFO9wrmgyWVf0_Dh_n5Im7ACVEhVxtAnLppzJvk4/edit?usp=sharing

\section{End-to-End System for EAE}
\label{sec:end_to_end}
\subsection{System Architecture}

This section introduces our novel end-to-end EAE system, which efficiently extracts event-related information from text by seamlessly identifying entity boundaries, determining their types, and recognizing argument entities and their relations to an event entity. We utilized the EAE NLI model to construct the system. As illustrated in Figure \ref{fig:inference}, the overall steps are:

\noindent \textbf{Named Entity Recognition Module:} The system starts by extracting entities and their types from the input sentence. The process is carried out through the online Wojood web service\footnote{{\small \url{https://sina.birzeit.edu/wojood}}}. Then, the system evaluates the extracted entities. If an event entity is recognized in a sentence, then other entities in this sentence are considered candidate arguments for this event.  

\noindent\textbf{Template Selection Module}: For each entity, a template is selected based on its category and used to construct the hypothesis. 

\noindent\textbf{Argument Extraction Module:} The input sentence is paired with the template and sent to the EAE NLI model to identify the argument entities for an event and their corresponding relationships. The type of template serves as the basis for establishing the relationship between the event and the entity. Specifically, if the EAE NLI model indicates a positive connection between the input sentence and the template, then the entity and the event linked within the template are considered to have a relationship indicated by the template label.

\subsection{Computing EAE baselines} 
The performance of our proposed EAE system is evaluated using \hadathNLI and \testNLI test sets. Results are shown in Table \ref{table:end_to_end_results}. Note that the relation classifiers share parameters, and the evaluation assumes named entities to be correctly recognized.

\begin{table}[h!]
\small
\centering
\begin{tabular}{|l|rrr|r|}
\hline
\textbf{Dataset} & \textbf{$P$} & \textbf{$R$} & \textbf{$F_1$} \\
\hline
\hadathNLI&93.45 & 94.52 &93.99 \\
\testNLI &  67.79 & 83.68 & 74.90  \\
\hline
\end{tabular}
\caption{ \label{table:end_to_end_results} Baselines: evaluation of our EAE system.}
\end{table}

\section{Ablation Studeis}
\label{sec:ablation_study}
\noindent\textbf{Best Template}: to choose the best template to implement in the EAE system for each relation (Figure \ref{tab:relation-templates_train}), we evaluated them using the test set. Table \ref{table:template_ablation_study} shows that $t_2$ performed slightly better.
\begin{table}[h!]
\small
\centering
\begin{tabular}{cccc}%{{p{1.8cm} p{1.5cm}p{0.8cm}p{0.8cm}p{0.8cm}}
\hline
Template & \textbf{$P.$} & \textbf{$R.$} & \textbf{$F1.$}  \\
\hline
$t_{1}$ & 92.43 & 92.65 & 92.54 \\ 
\textbf{$t_{2}$} & 92.63 & 92.60 & \textbf{92.61} \\
$t_{3}$ & 92.01 & 92.83 & 92.40 \\
$t_{4}$ & 92.35 & 92.40 & 92.37 \\
\hline
\end{tabular}
\caption{\label{table:template_ablation_study} Ablation study to choose the best template.} 
\end{table}

 \noindent\textbf{Best loss function}: We compared two loss functions: cross-entropy loss ($L_{CE}$) and Noise Contrastive Estimation Loss ($L_{NCE}$).  Table \ref{table:ablation_results} shows that while $L_{NCE}$ slightly outperforms $L_{CE}$ on the \hadathNLI, its performance significantly improves on \testNLI, showing its efficacy across diverse contexts.

Additionally, we improved the model's performance by combining weighted cross-entropy loss ($L_{WCE}$) with $L_{NCE}$, using weights $w_p=1$ and $w_n=0.5$. This approach, with higher weights for positive relations and slightly lower weights for negative ones, yielded the best results.
\begin{table} [ht!]
\small
\centering
\begin{tabular}{p{1.4cm} p{.8cm} p{1.1cm} p{0.75cm} p{0.75cm} p{0.75cm}}
\hline
\textbf{Loss Fn.}&\textbf{Class}& \textbf{Support} & \textbf{$P$} & \textbf{$R$} & \textbf{$F1$} \\
\hline
Cross &Neg.&1494 &  94.94 & 87.18 & 90.89  \\
Entropy&Pos. & 387& 62.94 & 82.43 &71.38  \\
\small($L_{CE}$)&Avg. & &  & & 81.14 \\
\hline
$Loss$&Neg.&1494 &  95.88 & 90.29 & 93.00  \\
\small($w_p=1$,&Pos. & 387& 69.41 & 85.01 &76.42  \\
\small$w_n=1$)&Avg. & &  & & 84.71 \\
\hline
$Loss$&Neg.&1494 &  94.44 & 92.10 & 93.26  \\
\small($w_p=1$,&Pos. & 387& 72.17 & 79.07 &75.46  \\
\small$w_n=.2$)&Avg. & &  & & 84.36 \\
\hline
$Loss$&Neg.&1494 &  95.75 & 90.56 & 93.09  \\
\small($w_p=1$,&Pos. & 387& 69.87 & 84.50 &76.49  \\
\small$w_n=.5$)&Avg. & &  & & \textbf{84.79} \\
\hline
\end{tabular}
\caption{ \label{table:ablation_results} Results (F1-score \%) on \hadathNLI test set.}
\end{table}

\section{Conclusion and Future work}
\label{sec:conclusion}
Our introduction of the \hadath and \testNLI corpora significantly advances Arabic event-argument extraction by providing a rich dataset with high inter-annotator agreement. Our novel BERT-based method for event relation extraction demonstrates exceptional performance, achieving high $F_1$ scores on both the \hadathNLI dataset and on \testNLI dataset. Additionally, our implementation of the EAE end-to-end system as part of the open-source SinaTools will enrich the Arabic NLP industry.
  
Large language models (LLMs) can further enhance our work by extracting information, including named entities and relationships, from text to populate knowledge graphs and improve other knowledge graph tasks like embedding and completion \cite{BCGJMSSV24}. In future work, we will explore the capabilities of LLMs to enhance event-argument extraction. Integrating LLMs into our framework could potentially improve the accuracy and scalability of our event extraction system.

\section*{Limitations}\label{sec:limits} 

The constructed \hadath and \testNLI corpora primarily focus on MSA data and do not cover dialectal variations. Furthermore, even though we included out-of-domain tests to assess performance, our results are constrained to the specific domains used in our study.

\section*{Ethics Statement}\label{sec:Ethics} 
The corpora provided for this research are derived from public sources, eliminating specific privacy concerns. The results of our research will be made publicly available to enable the research community to build upon them for the public good and peaceful purposes. Our data, tools, and ideas are strictly intended for non-malicious, peaceful, and non-military purposes.

\section*{Acknowledgements}
This research is partially funded by the research committee at Birzeit University. We extend our gratitude to Taymaa Hammouda for the technical support and to the students who helped and supported us during the annotation process, especially Haneen Liqreina and Sana Ghanim.

\bibliography{MyReferences,custom}
\end{document}